\documentclass[runningheads]{llncs}
\usepackage{amsmath}
\usepackage{graphicx}
\usepackage{fancyhdr}
\usepackage{atbegshi}
\usepackage{caption}
\usepackage{booktabs}
\usepackage{array}
\usepackage{multirow}
\usepackage{wrapfig}
\usepackage{algorithm}
\usepackage{algpseudocode}
\usepackage{float}
\usepackage{threeparttable, tablefootnote}
\usepackage{tabularx}
\usepackage{listings}
\usepackage{enumitem}
\begin{document}
	\title{Extracting Breast Cancer Phenotypes from Clinical Notes: Comparing LLMs with Classical Ontology Methods} 
	\titlerunning{}
	\author{Abdullah Bin Faiz\inst{1}, Arbaz Khan Shehzad\inst{1}, Asad Afzal\inst{1}, Momin Tariq\inst{1}, Muhammad Siddiqi\inst{1}, Muhammad Usamah Shahid\inst{1}, Maryam Noor Awan\inst{2}\thanks{Maryam Noor Awan is a lead of the expert panel of oncologists who evaluated the responses from the LLM.} \and Muddassar Farooq\inst{1}\thanks{Each author contributed equally.}}
	\authorrunning{A. Faiz et al.}
	\institute{
		\textit{CureMD Research, 80 Pine Street, New York, 10005, USA} \and Oncology Department, Shifa International Hospital, 4 Pitras Bukhari Rd, H-8/4 H 8/4 H-8, Islamabad, Pakistan
	}
	
	\maketitle
    \vspace*{-1em}
	\begin{abstract}
		A significant amount of data held in Oncology Electronic Medical Records (EMRs) is contained in unstructured provider notes -- including but not limited to the chemotherapy (or cancer treatment) outcome, different biomarkers, the tumor's location, sizes, and growth patterns of a patient. The clinical studies show that the majority of oncologists are comfortable providing these valuable insights in their notes in a natural language rather than the relevant structured fields of an EMR. The major contribution of this research is to report an LLM-based framework to process provider notes and extract valuable medical knowledge and phenotype mentioned above, with a focus on the domain of oncology. In this paper, we focus on extracting phenotypes related to breast cancer using our LLM framework, and then compare its performance with earlier works that used knowledge-driven annotation system, paired with the NCIt Ontology Annotator. The results of the study show that an LLM-based information extraction framework can be easily adapted to extract phenotypes with an accuracy that is comparable to the classical ontology-based methods. However, once trained, they could be easily fine-tuned to cater for other cancer types and diseases.
	\end{abstract}
	
	
\section{Introduction}
AI-centered electronic health applications are enabled by training on large numbers of patient records stored in EMR/EHR systems. These records usually contain medical history, lab and vital reports, diagnoses, and medications and their outcomes. These EMR systems -- containing structured, semi-structured and unstructured data -- coupled with advanced AI/ML models are at the core of Real-World Evidence (RWE) paradigm. Earlier studies show that oncologists prefer noting information related to treatment outcomes and other information elements about a patient in their clinical notes due to the profession's legacy. As a consequence, valuable medical information, such as phenotypes and cancer treatment outcomes are buried in unstructured notes, as oncologists rarely record them in the structured fields of oncology EMRs. For example, in our partner oncology EMR, 97\% of oncologists record the phenotypes and treatment outcomes in their clinical notes and only 3\% record them in the relevant structured fields.  
	
In order to extract phenotypes from provider notes, the notes need to be annotated by expert oncologists. But their time and associated cost are making it economically infeasible to reliably annotate notes. As a result, valuable RWE cannot be built about the true outcomes of NCCN compliant cancer treatments, which are administered to the patients. Consequently, the value-based care in the domain of oncology (personalized to a patient's medical profile and history) may also not be achieved in real world because the treatments, suggested in the guidelines, are an outcome of clinical trials typically conducted at prestigious university hospitals for which less than 5\% of the population becomes eligible. Luckily, oncology EMR contains the treatment outcomes in the real world on real patients; as a result, we could extract them from notes and generate RWE showing the efficacy and effectiveness of NCCN treatment guidelines. Subsequently, we can provide RWE as the missing link to close the feedback loop, and transform the existing one-size-fits-all treatment paradigm to an efficient, effective and economical value-based personalized cancer care system.

As mentioned above, a smart and autonomous notes annotation engine is the first step towards a value-based healthcare system. In this paper, we will focus on using an LLM framework to annotate oncologist notes and extract phenotypes for breast cancer, and compare it with an earlier knowledge-driven system that uses the NCIt Ontology Annotator to annotate notes. The major contributions of this paper are: (1) the improvement of the existing ontology system; (2) the implementation of a Retrieval-Augmented Generation (RAG) method for retrieving information; (3) the introduction of an LLM-based system that extracts the required information; and (4) an in-depth comparison of LLMs performing information extraction compared to knowledge-driven approaches.
	
\section{Related Work}
D'Souza and Ng \cite{d2015sieve} used a sieve-based approach to simply normalize disorder mentions in notes, and they were unable to identify annotation candidates in oncologist notes.
	
For medical text annotation, National Cancer Institute thesaurus (NCIt) \cite{golbeck2003national} provides stable and unique codes for over 171,000 classes and 500,000 relationships. The Cancer Care Treatment Outcome Ontology (CCTOO) \cite{lin2018cancer} consists of a total of 1,133 classes.
	
Named Entity Disambiguation (NED) leverages the power of medical ontologies for medical text annotation. In \cite{wang2015language}, the authors link entities by using entropy-based weights for knowledge bases to generate and rank domain-independent candidate graphs. Zheng et al. \cite{zheng2015entity} utilize a knowledge base of 300 biomedical ontologies to retrieve candidates, which are then used to construct document graphs to detect entities and entity links. 

Deep learning models such as BioBERT \cite{lee2020biobert} and SciBERT \cite{beltagy2019scibert} have F1 scores of over 90\% across various entity recognition tasks, but are limited in the types of entities they can detect. The authors in \cite{simmons2024benchmarking} evaluated several LLMs -- GPT 3.5 and GPT 4 -- to extract ICD-10-CM codes from clinical notes. The conclusion, however is that LLMs perform poorly in the extraction of ICD-10-CM when compared with the performance of a human coder.

In comparison, Dagdelen et al. \cite{dunn2022structured} trained GPT 3 with hundreds of prompt pairs, and the resultant GPT 3 converts information in medical notes into structured formats such as JSON. Huang et al. \cite{huang2024critical} used GPT 3.5 Turbo to extract information from pathological reports, identifying key features of lung and bone cancers. Despite some challenges with specialized medical terminology, this work shows that LLMs could be re-engineered and fine-tuned for extracting information from provider notes with minimum human supervision.

This paper focuses on using LLMs for information extraction from provider notes. The LLM-based system utilizes an LLM paired with knowledge-driven techniques to build a robust system capable of moving the phenotypes, extracted by the LLMs, and storing them in structured data for uses cases such as storing in structured fields of EMR, and creating training, testing and validation data sets for AI systems.
	
\section{Methodology}

\begin{figure}[H]
    \vspace*{-2em}
    \centering
    \includegraphics[width=1\textwidth]{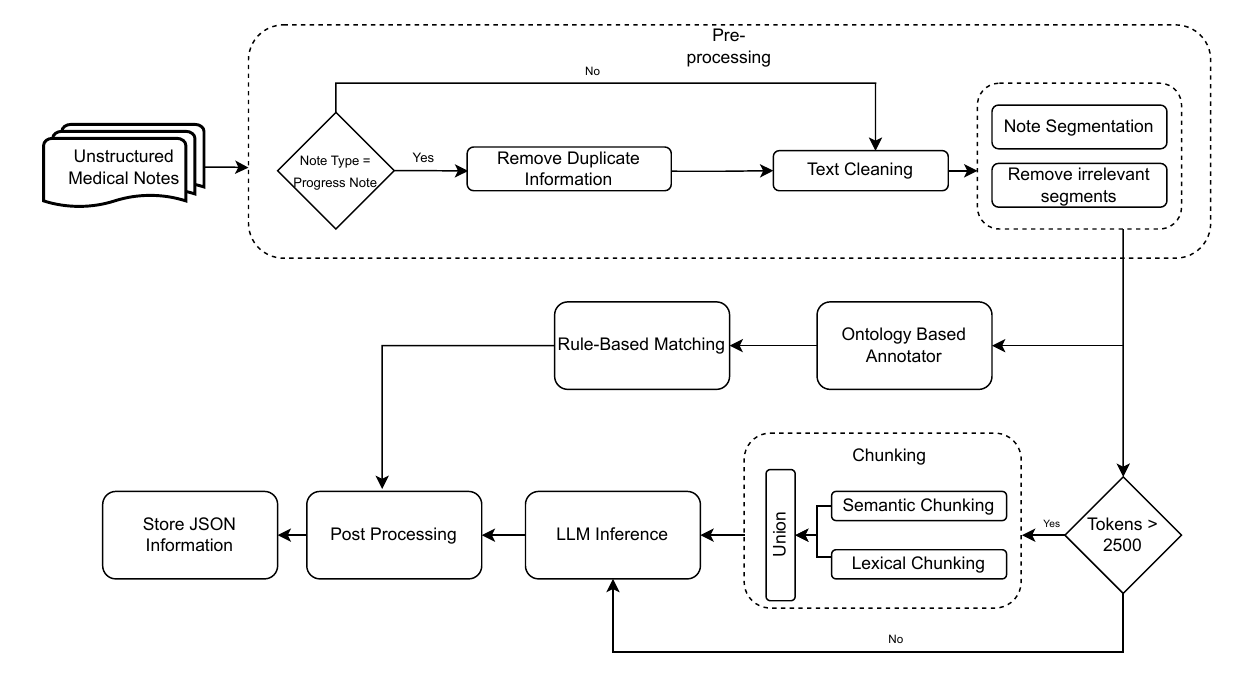}
    \caption{Complete pipeline to extract JSON information from unstructured text.}
    \label{fig:pipeline}
    \vspace*{-2em}
\end{figure}

The overall pipeline of our LLM Annotator framework starts with the pre-processing of clinical notes. The pre-processed notes are sent to the LLMs. In this study, we used and evaluated two well-known open-source LLMs for annotations: MistralAI's Mistral v0.2 and Meta's LLaMA 3 8B. Due to security and privacy concerns, the providers did not allow transmitting medical notes to cloud-based LLMs like GPT 4. They required on-premises and on-device inference.

\subsection{Pre-processing Pipeline}
The pre-processing pipeline first removes information elements that may not be relevant to breast cancer phenotypes, such as physical examinations, current medications, vital readings, allergies and Non-ASCII tags, to reduce token length. Moreover, we also removed redundant information about a patient in successive provider notes. The \textbf{differential text}, a note containing unique information, is obtained as described in Algorithm \ref{algo:PDT}.

\vspace*{-1em}
\begin{algorithm}[H]
    \caption{Process Differential Text. $df$ contains the patients' data, and $rmDups$ checks for duplication of notes information.}
    \begin{algorithmic}[1]
        \For{$r = 0$ \textbf{to} length of $df$}
        \If{$r = 0$ \textbf{or} $has\_changed$ is True}
        \State $df[r, 'diff\_text'] \gets df[r, 'note\_text']$
        \Else
        \State $df[r, 'diff\_text'] \gets rmDups(df[r-1, 'note\_text'], df[r, 'note\_text'])$
        \If{$df[r, 'diff\_text'] = df[r, 'note\_text']$}
        \State $df[r, 'diff\_text'] \gets rmDups(df[r-2, 'note\_text'], df[r, 'note\_text'])$
        \EndIf
        \EndIf
        \EndFor
    \end{algorithmic}
    \label{algo:PDT}
\end{algorithm}
\vspace*{-1em}
	
\noindent{As of now, the LLM-based framework has processed more than 95,000 breast cancer provider notes and stored the associated annotations in our secure on-premises lakehouse.}

\subsection{System 1 - Ontology-Based System}
The Smart Ontology-Based Annotator is a robust system developed using NLP techniques to extract phenotypes from oncologists' notes.\footnote{This system was presented in the SEPDA workshop of AIME 2023 proceedings.} This system annotates the notes using NCIt ontology and extracts important phenotypes from annotated notes, namely TNM staging, stage group, tumor size, cancer grade and performance, and biomarkers. The annotator identifies and links potential biomedical concepts in the text using NCIt and other biomedical ontologies.

We updated the system to extract information with improvements in accuracy. We also added the support for extracting Biomarkers: Most practices label estrogen (ER), progesterone (PR), and human epidermal growth factor receptor 2 (HER-2) with either proper keywords (positive, negative), abbreviations (pos., neg., etc.), or symbols (+, -). The regular expressions in the system were updated to better reflect the different notations and meanings.

A similar issue was identified for metastatic cancer patients. For some patients, either the metastasis is mentioned for another type of diagnosed cancer or is referred for someone related to the patient (e.g. mother, father etc.) which makes it difficult to relate it with the patient. We eventually extracted a flag pertaining to metastatic cancer of the patient, especially if it refers to breast cancer. 

Despite the improvements made to the system, the ontology system suffers from drawbacks: (1) it fails to pick semantics within the note; (2) it does not generalize to additional phenotypes and diseases; (3) it takes a longer time to process the note than an LLM; and (4) the system is dependent on the availability of NCBO API.

\subsection{System 2 - LLM-Based System}
The LLM-based system, on the other hand, solves the problems faced by the ontology-based system. The system uses an in-house cluster of GPUs, providing security and privacy, and also removing any dependencies on external tools. LLMs specialize in picking up semantics within the presented information, can adapt and generalize well to additional diseases and phenotypes, and with the assistance of acceleration tools like Nvidia's TensorRT-LLM, processes notes significantly faster than the ontology-based system.

The LLM is given an instruction through prompt engineering and a JSON schema that contains the same phenotypes as that of the ontology-based system. After the response is retrieved, the output is validated for any errors including incomplete JSON objects and enums.

\subsubsection{\textbf{Step 1. JSON Schema Designing.}}
\texttt{jsonschema}, a Python package, is used to create the JSON schema which allows us to set the key names, properties, types, enums, and patterns in the JSON. This schema acts as a blueprint for extracting and validating the structured information extracted from unstructured provider notes.
 
\subsubsection{\textbf{Step 2. Retrieval Augmented Generation (RAG).}}

\begin{algorithm}[!t]
    \caption{RAG Implementation}
    \begin{algorithmic}[1]
        \State $tokens \gets \text{tokenize}(note\_text)$
        \If {$\text{length}(tokens) > 2500$}
            \State $splitted\_texts \gets \text{recursive\_text\_splitter}(main\_text)$
            \State $embeddings \gets \text{create\_embeddings}(queries, splitted\_texts)$
            \State $cosine\_similarities \gets \text{calculate\_cosine\_similarity}(embeddings)$
            \State $top\_docs\_cosine \gets \text{select\_top\_documents}(cosine\_similarities, documents, k)$
            \State $indexed\_scores \gets \text{index\_documents}(queries, splitted\_texts)$
            \State $top\_docs\_index \gets \text{select\_top\_documents}(indexed\_scores, documents, k)$
            \State $final\_documents \gets \text{union}(top\_docs_cosine, top\_docs_index)$
            \State $ordered\_chunks \gets []$
            \For{each $doc$ in $original\_order$}
                \If{$doc$ in $final\_documents$}
                    \State $ordered\_chunks.append(doc)$
                \EndIf
            \EndFor
            \State $joined\_text \gets \text{concatenate}(ordered\_chunks)$
        \Else
            \State $joined\_text \gets note\_text$
        \EndIf
        
        \State $prompts \gets \text{create\_prompts}(queries, joined\_text, 2500)$
        \State $json\_outputs \gets \text{generate\_jsons}(prompts)$

        \Function{select\_top\_documents}{scores, documents, k}
            \State $scored\_documents \gets \text{zip}(scores, documents)$
            \State \text{Sort} $scored\_documents$ \text{by scores in descending order}
            \State $top\_documents \gets \text{list of first k documents from} \, scored\_documents$
            \State \Return $top\_documents$
        \EndFunction
    \end{algorithmic}
    \label{algo:RAG}
\end{algorithm}

Each LLM model that is evaluated has a context window defined by the token limit. We have to ensure that this token limit is not exceeded in any iteration of the extraction pipeline. Based on the experiments and the available information, we concluded that the notes with a token length of greater than 2500 generally exceed the token limit. Please see the distribution of token lengths in Appendix \ref{appx:token_lengths}. For these notes, we divided them into smaller chunks. The following steps are done for our customized RAG pipeline as described in Algorithm \ref{algo:RAG}.
	
\begin{enumerate}
	\item{\textbf{Chunking.}} We have leveraged Langchain's Recursive Chunking technique. Let a provider note be \(N\) and the chunks generated by the Recursive Text Splitter as \(C=\{c_1, c_2, \ldots c_n\}\). 
		
	\item{\textbf{Re-ranking.}} Once we have a set of chunks, we retrieve top \(k\) chunks with respect to our query prompt. We use a hybrid approach by modifying the method mentioned in \cite{wornow2024zero} and combining Lexical and semantic re-ranking.
    \begin{enumerate}
        \item{\textbf{Semantic Re-ranking.}} First, we create embeddings of query \(Q\) -- denoted by \(q_e\) -- and each chunk from the set of chunks \(C\) -- denote these embeddings as \(S = \{v_1, v_2, v_3, \ldots, v_n\}\) -- using \texttt{mxbai-embed-large-v1}. The cosine-similarity of each embedding \(s_i\), denoted as \(CosSim(s_i)\), is calculated with \(q_e\). The chunks are re-ordered based on the cosine-similarity scores. Let the new set be \(V'_n = \{v'_1, v'_2, v'_3, \ldots, v'_n\}\), and top \(k\) chunks are selected. In our case, we have used \(k=10\). We denote this new vector as \(C'_{10} = \{c'_1, c'_2, \ldots, c'_{10}\}\).
        \item{\textbf{Lexical Re-ranking.}} We use Pyserini's Lucene searcher which can efficiently score the chunks. Using the \texttt{BM25} algorithm, we compute the relevance scores of each chunk \(c_i\) and the query based on the frequency distribution of query terms within the \(c_i\). Once we have received all scores \(S = \{s_1, s_2, s_3, \ldots, s_n\}\), we reorder them to create a new set and select the top k chunks. We chose \(k=10\) for our case. Let \(S' = \{s'_1, s'_2, s'_3, \ldots, s'_n\}\) be the set of re-ordered chunks based on the scores. At the end, we select the top k most relevant chunks \(S'_{10} = \{s'_1, s'_2, s'_3, \ldots, s'_{10}\}\)
    \end{enumerate}

    Applying a union operation on both sets, we get the final set of chunks \(C_{finalized} = C'_{10}  \cup  S'_{10}\) that is sent to the LLM.
\end{enumerate}

\subsubsection{Step 3. Response Generation.} The chunks or the note, along with the JSON schema and prompt, are passed to the LLM using one-shot prompting, which contains medical information regarding the construction of the JSON object. Some examples of the prompt templates and samples queries are shown in Appendices \ref{appx:prompt_template} and \ref{appx:queries} respectively.

Once the LLM has produced the output, the generated JSON is extracted from the response. The JSON has to pass the following criteria: (1) the JSON must be valid, and (2) it must adhere to the provided schema. If any check fails, the JSON is passed iteratively to the LLM until the correct JSON is produced.

\subsubsection{Step 4. Post-processing Extracted JSON}

The JSON produced by LLMs is passed through a post-processing pipeline. This pipeline performs the following operations:
\begin{itemize}
    \item Standardizing dates, tumor size units, and boolean values by replacing them with values that follow the enums given in the schema.
    \item Fixing mismatched brackets in the generated response and correcting them to seamlessly store the JSON object in the lakehouse.
    \item Extracting values from key–value pairs and ensuring they conform to the schema.
\end{itemize}

\section{Results and Discussion}
\subsection{Results}
We created a sample of 150 randomly selected notes which comprises admission and progress notes drawn from randomly selected oncology practices to validate the results from both systems\footnote{We are grateful to CureMD for providing provider notes data of oncology patients, after anonymizing and de-identifying the data according to HIPAA guidelines, from its oncology EMR for this research.}. We extracted five labels from each note (one per phenotype), for a total of 750 labels. The notes were hand-annotated by a panel of five physicians headed by a senior resident oncologist in a partner university teaching hospital. Table \ref{tab:total_labels} shows the results gathered from different LLMs for different phenotypes.

\begin{table}[H]
    \vspace*{-1em}
    \begin{threeparttable}
    \caption{The Summary of results across different LLM Models for the phenotypes. Highest values are highlighted.}
    \begin{tabularx}{\textwidth}{l*{6}{>{\centering\arraybackslash}X}}
        \toprule
        \multirow{2}{*}{Model} & \multirow{2}{*}{Result} & \multicolumn{5}{c}{Phenotype} \\
        \cmidrule(lr){3-7}
        & & Biomarkers & Grade \& Perf. & Stage & TNM & Tumor \\
        \midrule
        \multirow{2}{*}{LLaMA 3 8B} & Correct & \textbf{133} & \textbf{140} & \textbf{124} & \textbf{123} & 126 \\
        & Incorrect & 17 & 10 & 26 & 27 & 24 \\
        \midrule
        \multirow{2}{*}{Mistral 7B} & Correct & 125 & 115 & 117 & 107 & \textbf{130} \\
        & Incorrect & 25 & 35 & 33 & 43 & 20 \\
        \midrule
        \multirow{2}{*}{Ontology*} & Correct & 82 & 102 & 121 & 107 & 111 \\
        & Incorrect & 41 & 21 & 2 & 16 & 12 \\
        \bottomrule
        \label{tab:total_labels}
    \end{tabularx}
    \vspace*{-3mm}
    \caption*{\textit{* - The total number sums up to 615 instead of 750. Out of 150 API calls, 27 failed to give a response when provided with the note.}}
    \end{threeparttable}
    \vspace*{-3em}
\end{table}

Some incorrect entries have reduced the precision and recall metrics given in Table \ref{tab:accuracy_metrics}. This is due to multiple reasons:
\begin{itemize}
    \item \textbf{Missing.} The system failed to pick a phenotype value, but it was present in the note.
    \item \textbf{Hallucination.} The system picked or made up a random value when it was not present in the note or was present with a different value.
    \item \textbf{No Response.} Exclusive to the ontology-based system, the API would sometimes fail to return a response for a given note, even when information was present in the note.
\end{itemize}

\begin{table}[h]
    \vspace*{-2em}
    \centering
    \caption{Accuracy metrics of the results. Since the ontology-based system does not hallucinate, its precision is reported to be 100\%.}
    \label{tab:accuracy_metrics}
    \begin{threeparttable}
    \begin{tabularx}{\textwidth}{l *{8}{>{\centering\arraybackslash}X}}
        \toprule
        \textbf{Model} & \textbf{Accuracy} & \textbf{Precision} & \textbf{Recall} & \textbf{F-1 Score} \\
        \midrule
        LLaMA 3 8B & 86.13\% & 87.90\% & 94.95\% & 91.29\% \\
        Mistral 7B & 79.20\% & 81.35\% & 91.98\% & 86.34\% \\
        Ontology & 85.04\% & 100.00\% & 83.80\% & 91.19\% \\
        \bottomrule
    \end{tabularx}
    \end{threeparttable}
\end{table}

For the ontology-based system, the experiments were conducted on an Intel Core i7 1255U @ 1.70 GHz with 24 Gigabytes of DDR4 RAM. For the LLM-based system, as mentioned, we used a cluster of 2 A100-40GB PCIe GPUs; hence, we could not deploy larger models of LLaMA 3 and Mixtral. Due to security and privacy concerns, the practices require that an LLM model be usable on Nvidia A100-type GPUs in an on-premises environment; therefore, an open source model like LLaMA 3 can provide the best of both worlds: (1) security and privacy of patients' data and inference models; and (2) comparable performance with that of state-of-the-art LLMs using fine-tuning options. 

Tables \ref{tab:total_labels} and \ref{tab:accuracy_metrics} show that Mistral's performance was inferior, although we believe that the larger variants of both Mistral and LLaMA 3 (8x7B and 70B respectively) would achieve better results. In comparison, LLaMA 3 achieved the best performance with an accuracy, precision, recall and F1-score of 86\%, 87\%, 94\%, and 91\% respectively. This shows that LLaMA 3 with our customized RAG was able to extract the majority of the phenotypes with correct values.

\subsection{Discussion}
Our analysis shows that LLMs, despite being black-box in nature, possess capabilities which are absent in classical ontology methods.
\begin{enumerate}
    \item \textbf{Captures Meaningful Information.} LLMs capture semantically relevant information with better accuracy.
    \item \textbf{Replication.} LLMs can be easily reconfigured to extract new phenotypes by updating the schema and prompts; while for ontology-based systems, the extraction algorithms need to be updated for such enhancements.
    
\end{enumerate}
The input parameters were kept the same across all LLMs for a fair and unbiased comparison (temperature, top\_k, top\_p, etc.)

Despite the limitations in covering new phenotypes, the ontology-based system still provides some benefits over LLMs. 
	\begin{enumerate}
        \item \textbf{Easier to implement.} Once phenotypes and patterns are identified, the implementation requires few lines of code that give reliable and repeatable performance with no hallucinations.
        \item \textbf{Light on resources.} The ontology-based systems are lightweight and can run on edge devices such as smartphones.
        \item \textbf{Robustness.} The ontology systems, therefore, are more robust but lack the ability to generalize.
    \end{enumerate}
The ontology-based system, on the average, processed notes in less than 20 seconds; while the LLM-based models, running on-premises on A100, processed the notes (on the average) in less than 12 seconds. 
      
\section{Conclusion}
We present in this paper two systems for extracting phenotypes from provider notes: (1) an ontology-based system; and (2) an LLM-based system. The results show that LLMs, once running on-premise, can provide two benefits: (1) security and privacy of patients' data and inference models; and (2) superior performance compared with ontology-based systems.

In the near future, we will have a cluster of 8 A100, and we will evaluate larger open source models like Mixtral 8x7B and LLaMA 3 70B. By fine tuning and optimizing the configurations of these larger models, we are confident that their performance would become comparable with that of proprietary cloud-based LLMs. With the consent of the providers and patients, we aim to use a subset of the notes for evaluating GPT-4 and Claude 3 as benchmarks and compare it to open-source LLMs. Last but not least, we also want to combine ontology-based system with LLMs to remove the problem of hallucinations in LLMs, yet provide a higher accuracy of phenotype extraction in real time and in a generalized manner. This will be the subject of forthcoming publications.
	
\bibliographystyle{splncs04}
\bibliography{refs}

\pagebreak

\appendix
\section{Token Length of Breast Cancer Notes}\label{appx:token_lengths}
	\begin{figure}
		\centering
		\includegraphics[width=0.8\textwidth]{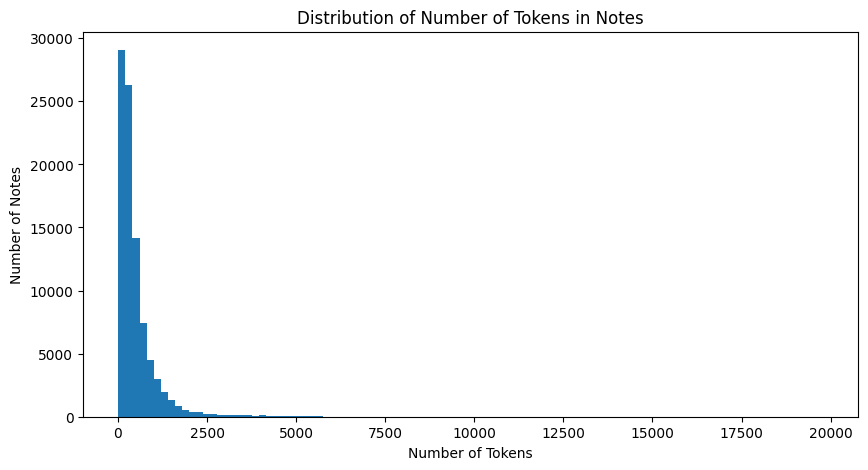}
		\captionsetup{justification=centering}
		\caption{Distribution of token lengths (in tokens) of the breast cancer notes, sampled across approximately 256,000 notes. Note that while the $x$ axis extends to 20000 and beyond with no visible marks, some notes exist with a relatively large number of tokens which cannot be fed directly to an LLM given its context window and the existing hardware constraints of our on-premises system.}
	\end{figure}
 \section{Prompt Template}\label{appx:prompt_template}
 
You are an expert in the Breast Cancer domain. You are also an expert in extracting information asked within a schema. Unfortunately, you can only respond in JSON format, but you make fantastic JSON objects free of errors. Your job is to, given a schema and a note context, extract the information from the notes that are required for the schema and make a JSON object that adheres to the schema guidelines. Here are some rules that you have to follow:

    - Do not return or modify the schema.
    - Do not make far-fetched assumptions. Obvious ones are fine.
    - If given enums, do not deviate from the ones provided.
    - Only provide the JSON object. Do not give anything else. Do not even add comments in the JSON object.
    - If anything specified in the schema is not present in the note, mark it as null. Do not remove it from the schema.
    - If there are multiple readings of the information to extract, pick the reading by the latest date.
Schema:
    \begin{lstlisting}
        ```json\n{Respective JSON Schema}\n```
    \end{lstlisting}
Example Text: 
    [The sample text of one shot example goes here.]
Example JSON:
    \begin{lstlisting}
    ```json\n{Example JSON extracted from example text.}\n```
    \end{lstlisting}
Actual Text: 
    [Actual Text from which information is to be extracted goes here.]
Query:
    [Actual query goes here]

\section{Queries}\label{appx:queries}
 \begin{lstlisting}
    TNM and Staging: 
        What is the stage of the cancer and the T, N and M values?
    Tumor Information: 
        List down all tumors found in the note above specifically tumor size, lesions, dates and masses.
    Grade and Performance: 
        What are the values for grade, ECOG, KARNOFSKY/KPS found in text above
    Biomarkers: 
        List HER-2, ER, PR  biomarkers that the patient was tested for and their values.
 \end{lstlisting}

\end{document}